\documentclass[]{fairmeta}

\usepackage{wrapfig}
\usepackage{tabularx}
\usepackage{textcomp}
\usepackage{stfloats}
\usepackage{url}
\usepackage{verbatim}
\usepackage{titlesec}
\usepackage{tocloft}
\usepackage{adjustbox}
\usepackage{multirow}
\usepackage{pifont}
\usepackage{tikz}
\usepackage{comment}
\usepackage{amsmath,amssymb} 
\usepackage{colortbl}  
\usepackage{color}
\usepackage{booktabs} 
\usepackage{hyperref}
\usepackage{graphicx}    
\usepackage{subcaption} 
\usepackage{multirow} 
\usepackage{booktabs} 
\usepackage{subcaption} 
\RequirePackage{xspace}
\makeatletter
\DeclareRobustCommand\onedot{\futurelet\@let@token\@onedot}
\def\@onedot{\ifx\@let@token.\else.\null\fi\xspace}

\makeatother

\preto\section{\FloatBarrier}

\definecolor{adptorange}{RGB}{248, 205, 172}
\definecolor{cmpblue}{RGB}{189, 215, 238}
\definecolor{cmpblue}{RGB}{189, 215, 238}

\definecolor{our_red}{RGB}{232,157,160}
\definecolor{our_blue}{RGB}{136,206,230}
\definecolor{our_orange}{RGB}{246,200,168}
\definecolor{our_green}{RGB}{178,211,164}

\definecolor{attn_code0}{RGB}{247,215,200}
\definecolor{attn_code1}{RGB}{238,169,139}
\definecolor{mlp_code0}{RGB}{204,201,221}
\definecolor{mlp_code1}{RGB}{102,95,153}

\definecolor{token_blue}{RGB}{84, 120, 140}

\usepackage{pifont}       
\usepackage{bbding}       
\usepackage{fontawesome}
\usepackage{xspace}

\usepackage{float}

\newlength\savewidth

\newcolumntype{x}[1]{>{\centering\arraybackslash}p{#1pt}}
\newcolumntype{y}[1]{>{\raggedright\arraybackslash}p{#1pt}}
\newcolumntype{z}[1]{>{\raggedleft\arraybackslash}p{#1pt}}

\renewcommand{\paragraph}[1]{\vspace{1mm}\noindent\textbf{#1}}
\usepackage{colortbl}
\usepackage{xcolor}
\usepackage{wrapfig}

\renewcommand{\paragraph}[1]{\vspace{1.25mm}\noindent\textbf{#1}}

\usepackage{algorithm}
\usepackage{listings}

\definecolor{codeblue}{rgb}{0.25, 0.5, 0.5}
\definecolor{codekw}{rgb}{0.35, 0.35, 0.75}
\lstdefinestyle{Pytorch}{
    language = Python,
    backgroundcolor = \color{white},
    basicstyle = \fontsize{9pt}{8pt}\selectfont\ttfamily\bfseries,
    columns = fullflexible,
    aboveskip=1pt,
    belowskip=1pt,
    breaklines = true,
    captionpos = b,
    commentstyle = \color{codeblue},
    keywordstyle = \color{codekw},
}

\definecolor{green}{HTML}{009000}
\definecolor{red}{HTML}{ea4335}

\setlabdisplayname{OmniAl Group of ZJU ACES Lab}
\setuniversityname{}

\title{GFT: From Imitation to Reward Fine-Tuning with Unbiased Group Advantages and Dynamic Coefficient Rectification}

\author[* 1]{Wangjie Gan}
\author[* 1]{Miao Pan}
\author[* 1]{Linbo Xi}
\author[\dagger 1]{Wenqi Zhang}
\author[1]{Jintao Chen}
\author[1]{Jianwei Yin}
\author[\dagger 1]{Xuhong Zhang}

\affiliation[1]{School of Software Technolog,Zhejiang University\\}
\contribution[*]{Equal contribution}
\contribution[\dagger]{Corresponding author}
\correspondence{\texttt{\{zhangwenqi,zhangxuhong\}@zju.edu.cn}}

\abstract{
Large language models are typically post-trained using supervised fine-tuning (SFT) and reinforcement learning (RL), yet effectively unifying efficient knowledge injection with robust generalization remains challenging. In this work, we provide a training-dynamics analysis showing that SFT can be interpreted as a special case of policy gradient optimization with an extremely sparse implicit reward and unstable inverse-probability weighting, which together lead to single-path dependency, entropy collapse, and gradient explosion. Motivated by this diagnosis, we propose Group Fine-Tuning (GFT), a unified post-training framework that addresses these intrinsic limitations through two mechanisms: Group Advantage Learning, which constructs diverse response groups and derives normalized contrastive supervision to alleviate reward sparsity, and Dynamic Coefficient Rectification, which adaptively bounds inverse-probability weights to stabilize optimization while preserving efficient knowledge injection. Experiments demonstrate that GFT consistently surpasses SFT-based methods and yields policies that integrate more smoothly with subsequent RL training.
}

\date{April 2026}

\begin{document}
\thispagestyle{firstheader}
\maketitle
\pagestyle{empty}

\section{Introduction}

The remarkable advancement of large language models has been driven to a great extent by two core post-training techniques: supervised fine-tuning (SFT) and reinforcement learning (RL)~\cite{guo2025deepseek, xu2025qwen2}.
A substantial body of prior work has investigated the respective strengths of these two paradigms.
SFT leverages expert demonstration data to efficiently inject knowledge and skills, enabling models to rapidly acquire instruction-following abilities and domain-specific competence~\cite{chu2025sft, chung2024scaling}.
Meanwhile, RL guides models to explore and optimize within a broad policy space through reward signals, facilitating the learning of robust reasoning behaviors and generalizable strategies~\cite{guo2025deepseek,wang2024math}.



\begin{figure*}[h]
    \centering
    \includegraphics[width=1\linewidth]{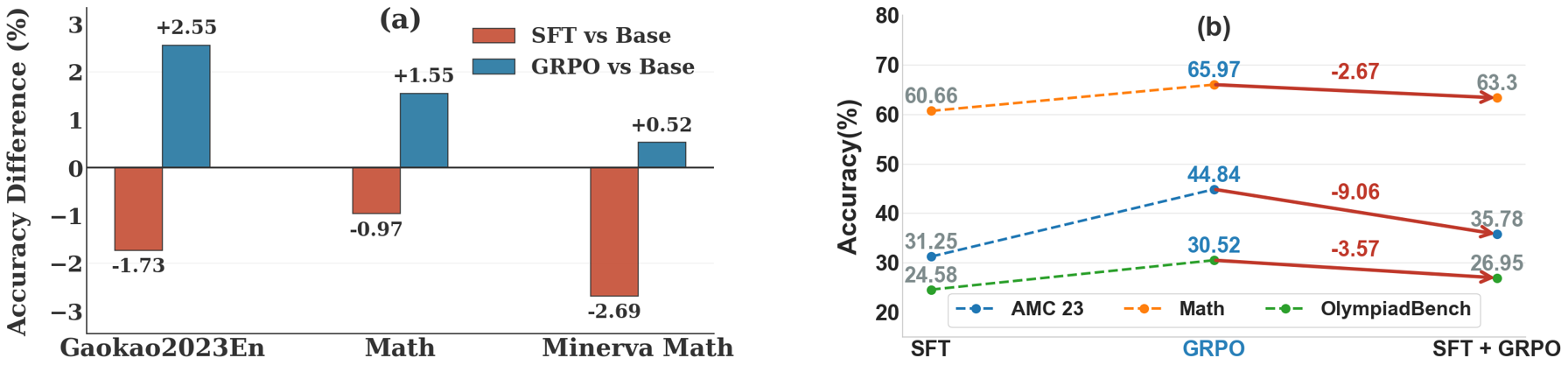}
    \caption{Performance of Qwen2.5-Math-1.5B on Numina-Math. (a) Accuracy changes relative to the base model: SFT consistently degrades performance, highlighting catastrophic forgetting. (b) Accuracy across different training pipelines: the SFT+GRPO pipeline exhibits poor synergy, underperforming GRPO alone.}
    \label{fig:intro}
\end{figure*}

Despite the complementary strengths of SFT and RL, SFT training is highly sensitive to high-fidelity expert data~\citep{Zhou_Liu_Ai,gudibande2023false} and often exhibits unstable optimization, which manifests in two salient failure modes in Figure \ref{fig:intro}. First, the strict imitation objective can overwrite and shift general-purpose representations acquired during pretraining, leading to catastrophic forgetting~\citep{aw2023instruction,chu2025sft,ruan2025unveiling,luo2025empirical} and degraded out-of-distribution generalization—consistent with the systematic regressions of SFT relative to the Base model in Figure \ref{fig:intro}(a). Second, SFT tends to over-constrain the policy to a narrow demonstration manifold, reducing policy entropy and solution diversity and thereby shrinking the exploration budget required by downstream RL~\citep{chen2025sft,chen2025synergy,qin2025supervised}; as a result, Figure \ref{fig:intro}(b) shows a clear synergy break where RL alone (e.g., GRPO) delivers substantial gains, yet the common sequential pipeline (SFT+GRPO) yields consistently diminished improvements, i.e., “RL works, but its benefits are attenuated when preceded by SFT.”


To investigate the root causes of these challenges, we present a principled theoretical analysis from the perspective of training dynamics. We demonstrate that SFT can be interpreted as a special case of reinforcement learning, but one that suffers from two fundamental flaws: (1) It is constrained by \textbf{single-path dependency}, where the implicit reward $r(x,y)=\mathbb{I}[y=y^{*}]$ restricts the learning signal to the exact expert trajectory, leading to \textbf{insufficient exploration} and \textbf{entropy collapse}. (2) It is vulnerable to \textbf{gradient explosion} during optimization. Since the gradient updates are scaled by an \textbf{unstable importance weight} $w(y|x)=1/\pi_{\theta}(y|x)$ (the reciprocal of the token probability), valid but unfamiliar expert tokens cause this weight to grow excessively large, triggering \textbf{gradient explosion} and driving the model toward \textbf{mechanical memorization} and \textbf{overfitting}. Together, these factors constitute the mathematical explanation for SFT's limited generalization ability.

Motivated by these theoretical insights, we propose \textbf{Group Fine-Tuning (GFT)}, a unified post-training paradigm designed to directly mitigate these intrinsic deficiencies.
GFT introduces two key mechanisms.
\emph{Group Advantage Learning} overcomes SFT's single-path dependency by creating a diverse response group for each query, combining model-generated samples, expert demonstrations, and teacher outputs.
By evaluating candidates according to their normalized within-group advantages, rather than rigidly imitating expert data, this approach produces learning signals that are comparable across diverse responses, thereby preserving essential exploration during early post-training.
\emph{Dynamic Coefficient Rectification} stabilizes optimization while preserving learning capacity through a clipping-like adaptive weighting scheme.
By applying a dynamic threshold $\tau$ to the \textbf{importance weight} $w(y|x)$, this mechanism suppresses gradient explosion for extreme samples while \textbf{preserving the effective gradient} for moderately low-probability tokens, enabling efficient injection of new knowledge into models.


We systematically evaluated GFT across multiple model families and math-reasoning benchmarks. Compared with standard SFT, strong SFT variants such as DFT \cite{wu2025generalization} and ASFT \cite{zhu2025anchored}, RL baselines such as GRPO, and component-wise ablations, GFT consistently outperforms all baselines on both standard and competition-level tasks with substantially higher data efficiency. To further probe the post-training “synergy dilemma,” we use GFT as the initialization for subsequent RL and contrast it with the conventional “SFT→RL” pipeline; GFT provides a stronger cold start and more stable optimization, thereby significantly raising the attainable performance ceiling of RL. Finally, evaluations of catastrophic forgetting and output diversity show that GFT markedly mitigates the severe forgetting typical of SFT while achieving a practical unification of improved precision and preserved exploration.

\noindent\textbf{Our main contributions include:}
\newline\textbullet\ From a training-dynamics perspective, we identify two causes of SFT’s weak generalization:
(i) inherent single-path dependency, where each context is supervised by a single expert demonstration; and
(ii) gradient explosion, which promotes mechanical memorization and catastrophic forgetting.
\newline\textbullet\ We propose \textbf{GFT}, unifying unbiased group advantages and token-wise update stabilization into a single-stage post-training procedure by combining group advantage learning with dynamic importance weight rectification.
\newline\textbullet\ Extensive experiments across multiple benchmarks show that GFT consistently outperforms standard SFT and strong SFT-based baselines, validating GFT as a foundational post-training paradigm for LLMs.

\section{Preliminaries}\label{sec:preliminaries}

In SFT learning process, the policy model $\pi_{\theta}$ is trained to imitate expert demonstrations. Given a expert dataset $\mathcal{D}=\{(x,y^{*})\}$, the gradient of the SFT objective with model parameters $\theta$ is
\begin{equation}
    \nabla_{\theta}\mathcal{L}_{\mathrm{SFT}}
    =
    \mathbb{E}_{\mathcal{D}}
    \big[
    -\nabla_{\theta}\log \pi_{\theta}(y^{*}\mid x)
    \big].
    \label{eq:sft_grad}
\end{equation}

This gradient increases the likelihood of the expert-provided response and does not explicitly consider alternative outputs. However, in RL training process, the output $y$ is generated by the current model $\pi_{\theta}(\cdot\mid x)$ itself. The reward $r(x, y)$ is then computed for this model-generated sample. The policy gradient takes the form
\begin{equation}
    \nabla_{\theta}\mathcal{L}_{\mathrm{RL}}
    =
    \mathbb{E}_{x,y}
    \big[
    -\nabla_{\theta}\log \pi_{\theta}(y\mid x)\, r(x,y)
    \big].
    \label{eq:rl_grad}
\end{equation}

Notably, SFT can be viewed as a special case of RL. Specifically, if we interpret the SFT objective as maximizing a sparse reward that only provides non-zero feedback for the expert trajectory, its gradient can be rewritten as an on-policy expectation over $\pi_{\theta}$ via importance sampling. This equivalent formulation is:
\begin{equation}
\begin{split}
    \nabla_{\theta}\mathcal{L}
    =
    -\mathbb{E}_{x, y}
    \Big[
    \frac{\mathbb{I}[y=y^{*}]}{\pi_{\theta}(y\mid x)}
    \nabla_{\theta}\log \pi_{\theta}(y\mid x)
    \Big],
\end{split}
\label{eq:sft_final_form}
\end{equation}
where the indicator $\mathbb{I}[y=y^{*}]$ serves as a sparse reward that assigns a unit signal only when the sampled output exactly matches the expert demonstration, and zero otherwise. The term $1/\pi_{\theta}(y\mid x)$ corresponds to an importance weight that corrects for sampling from the current policy $\pi_{\theta}$ instead of the expert distribution.
The detailed derivation is provided in Appendix \ref{sec:proof}.

Eq.~\eqref{eq:sft_final_form} exposes two intrinsic limitations of SFT from an RL perspective. First, \textbf{single-path dependency} arises as the sparse reward confines learning to a single expert trajectory, offering no comparative feedback over alternatives. Second, \textbf{gradient explosion} occurs because the importance weight $1/\pi_{\theta}(y\mid x)$ grows excessively when the expert action probability is small, leading to highly unstable optimization behavior.

\section{Method: Group Fine Tuning}

\begin{figure*}[h]
    \centering
    \includegraphics[width=1\linewidth]{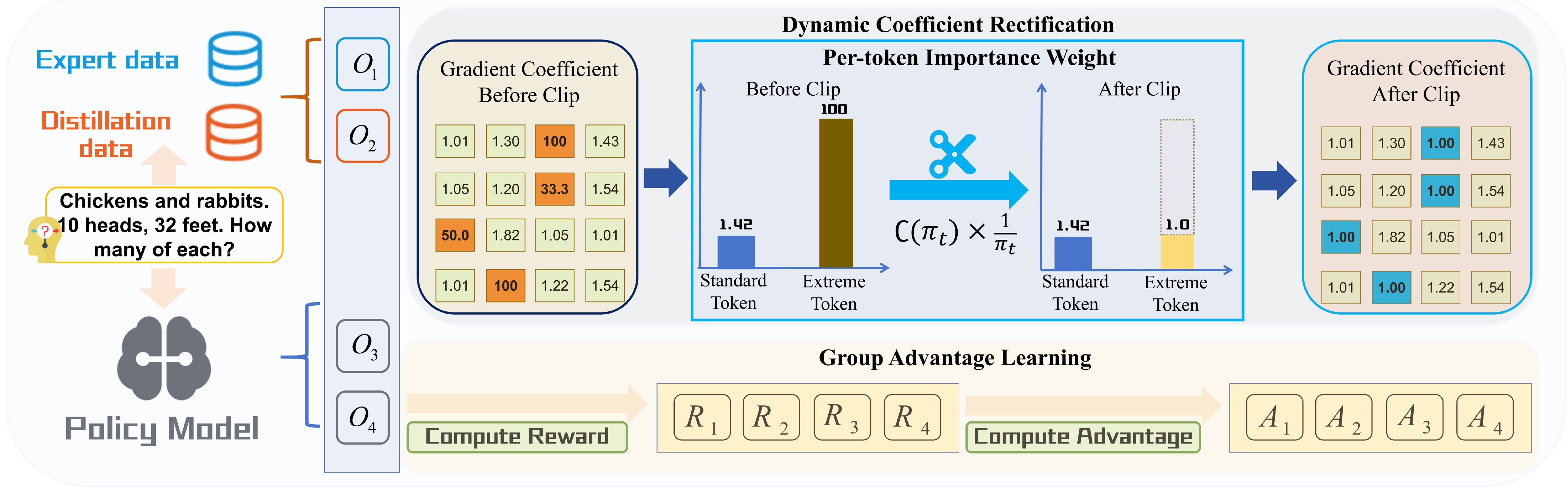}
    \caption{GFT comprises two components: (1) \textit{Group Advantage Learning}, which computes standardized relative advantages ($A_k$) from hybrid response groups (expert demonstrations, teacher outputs, and rollout samples); and (2) \textit{Dynamic Coefficient Rectification}, which bounds importance weights via per-token gradient clipping.}
    \label{fig:overview}
\end{figure*}

To address the intrinsic limitations of SFT identified in Eq.~\eqref{eq:sft_final_form}, we propose two complementary mechanisms.
\textbf{Group Advantage Learning} (GAL) constructs a group of multiple candidate trajectories for one query and evaluates each trajectory based on rule-consistent rewards, allowing the model to learn from diverse reasoning paths rather than treating only expert demonstrations as correct.
\textbf{Dynamic Coefficient Rectification} (DCR) stabilizes training by clipping the weight $1/\pi_{\theta}(y\mid x)$ for extremely low-probability tokens, preventing gradient explosion while preserving the original gradient for standard tokens to ensure efficient knowledge injection.

\subsection{Group Advantage Learning}

To move beyond the limitations of single-path dependency, we expand the standard SFT dataset into a comprehensive hybrid response group $\mathcal{G}x = {y_1, ..., y_K}$ for each query $x$. This group strategically integrates three complementary data sources: \textbf{Expert Demonstrations} ($y_{\text{exp}}$) that provide ground truth to guarantee a valid optimization direction always exists; \textbf{Teacher Distillations} ($y_{\text{demo}}$) from other powerful models, introducing diverse reasoning paradigms to break single-path dependency; and \textbf{Self-Generated Samples} ($y_{\text{sample}}$) obtained from the model's own rollouts, offering on-policy feedback to rectify intrinsic errors while reinforcing successful self-exploration. This design maintains high flexibility, allowing the composition to adapt based on data availability and training objectives. To effectively utilize the strengths of each data source within a unified learning framework, we assign a scalar reward $R(y_k)$ to each response in group $\mathcal{G}_x$, then compute a standardized advantage score:
\begin{equation}
A(y_k) = \frac{R(y_k) - \mu(\mathcal{G}_x)}{\sigma_R(\mathcal{G}_x) + \epsilon},
\label{eq:advantage}
\end{equation}
where \(\bar{R}(\mathcal{G}_x)\) and \(\sigma_R(\mathcal{G}_x)\) denote the mean and standard deviation of rewards within the group, and \(\epsilon > 0\) is a small constant that ensures numerical stability. This normalization centers and scales the rewards, creating a \textbf{relative, contrastive signal} within the group. Consequently, the reward mechanism guides the model to discern and prioritize high-quality responses, effectively unifying imitation, distillation, and self-improvement within a single, stable objective.

\subsection{Dynamic Coefficient Rectification}

The theoretical analysis in Eq.~\eqref{eq:sft_final_form} reveals that the inverse probability term $1/\pi_{\theta}$ introduces an inherent instability into the SFT-style optimization.
In practice, this instability arises in two common and complementary scenarios.
First, when the model increases its exploration by rolling out uncertain or diverse responses, the predicted token probabilities $\pi_t$ can become small, causing the corresponding update coefficients to grow excessively large.
Second, even when fitting expert demonstrations or teacher-distilled responses, the model may initially assign low probability to valid but unfamiliar tokens, which similarly amplifies the inverse weighting term. Inspired by the gradient clipping technique prevalent in RL, we propose a simple rectification function to stabilize the training:
\begin{equation}\label{eq:rectification_factor}
        \mathcal{C}(\pi_t) = 
        \begin{cases} 
            \text{sg}(\pi_t) & \text{if } \pi_t < \tau\\ 
            1 & \text{if } \pi_t \ge \tau
        \end{cases}
\end{equation}

Here, $\tau$ is a confidence threshold, and $\text{sg}(\cdot)$ denotes the stop-gradient operator. This design actively suppresses the explosive term $1/\pi_t$ for low-confidence tokens ($\pi_t < \tau$) by using $\text{sg}(\pi_t)$ to yield a bounded effective coefficient, while leaving the gradient unchanged for confident predictions ($\pi_t \ge \tau$). This ensures stable updates during exploration and preserves full learning strength for knowledge transfer, effectively resolving the instability inherent in the SFT objective.

\subsection{Final GFT Objective}
Combining Group Advantage Learning and Dynamic Coefficient Rectification, we derive the final training objective in its gradient form.
{\small
\begin{equation}\label{eq:gft_gradient}
\nabla_{\theta}\mathcal{L}
=
\mathbb{E}_{y_k {\in} \mathcal{G}_x}
\Big[
A(y_k)
\frac{\mathcal{C}(\pi)}{\pi_{\theta}(y_k|x)}
\nabla\!\log \pi_{\theta}(y_k|x)
\Big].
\end{equation}
}
Eq.~\eqref{eq:gft_gradient} presents the sequence-level gradient of GFT; the corresponding token-level formulation and loss definition are provided in Appendix \ref{sec:formulation}.
This gradient directly resolves the two intrinsic limitations of SFT: group-wise advantage weighting introduces contrastive supervision across multiple trajectories, while dynamic coefficient rectification bounds the update magnitude for low-probability tokens to prevent gradient explosion.



\section{Experiments}
\subsection{Experimental Setup}
\label{sec:experimental_setup}

\paragraph{Baselines and Models}
We compare GFT against a diverse set of paradigms, ranging from standard SFT and its recent stabilized variants—DFT~\citep{wu2025generalization}, ASFT~\citep{zhu2025anchored}, and PSFT~\citep{zhu2025proximal}—to the reinforcement learning baseline GRPO. Following DFT~\citep{wu2025generalization}, we evaluate five models covering diverse sizes, types and architectures: Qwen2.5-Math (1.5B, 7B)~\citep{yang2024qwen2}, LLaMA-3 (3.2-3B, 3.1-8B)~\citep{dubey2024llama}, and DeepSeekMath-7B-Base~\citep{shao2024deepseekmath}.

\paragraph{Training Settings}
Following prior works~\citep{wu2025generalization,zhu2025anchored,ming2025one,zhou2025variational}, we utilize the NuminaMath CoT dataset~\citep{numina_math_datasets}, selected for its extensive diversity ranging from high school exercises to international olympiads. For GFT, we construct a hybrid response group of size $K=8$ per query, comprising 1 expert demonstration, 3 teacher distillations from Qwen2.5-Math-72B, and 4 self-generated samples. Similarly, the GRPO baseline is configured to generate 8 outputs per query. To align total training volume, GFT and GRPO utilize a 10k subset (8 trajectories per query), whereas single-trajectory baselines (e.g., SFT) use 100k samples. See Appendix~\ref{sec:eval_settings} for evaluation details.


\subsection{Main Results}
\label{sec:main_results}

\begin{table*}[!t]
    \centering
    \small 
    \caption{Main results on seven math benchmarks. \textbf{SFT(mix)} indicates that the dataset is a mixture of expert datasets and distilled teacher datasets, while \textbf{GFT(no mix)} represents using only expert datasets without distilled data. \textbf{Bold} and \colorbox{blue!10}{blue} denote the best intra-group and overall performance, respectively. Overall, GFT achieves the best average performance across diverse model scales.}
    \label{tab:main_results}
      
    \resizebox{\textwidth}{!}{%
        \setlength{\tabcolsep}{3pt} 
        \renewcommand{\arraystretch}{1.2} 
        \begin{tabular}{llcccccc}
            \toprule
            \textbf{Model} & \textbf{Method} & \textbf{AMC23} & \textbf{College Math} & \textbf{Gaokao2023En} & \textbf{Math} & \textbf{Minerva Math} & \textbf{TabMWP} \\
            \midrule
              
            \multirow{9}{*}{\textbf{Qwen2.5-Math-1.5B}} 
            & Base Model & 30.16 & 24.30 & 34.81 & 46.54 & 10.51 & 24.55 \\
            & + SFT & 31.25 (+1.09) & 36.45 (+12.15) & 48.86 (+14.05) & 60.66 (+14.12) & 23.99 (+13.48) & 79.34 (+54.79) \\
            & + SFT(mix) & 32.70 (+2.54) & 36.35 (+12.05) & 50.82 (+16.01) & 60.41 (+13.87) & 25.76 (+15.25) & 80.15 (+55.60) \\
            & + GRPO & 44.84 (+14.68) & 35.58 (+11.28) & 51.80 (+16.99) & 65.97 (+19.43) & 21.17 (+10.66) & 76.94 (+52.39) \\
            & + ASFT & 43.12 (+12.96) & 29.40 (+5.10) & 47.99 (+13.18) & 60.35 (+13.81) & 15.55 (+5.04) & 65.06 (+40.51) \\
            & + PSFT & 31.56 (+1.40) & 33.77 (+9.47) & 47.66 (+12.85) & 59.51 (+12.97) & 19.13 (+8.62) & 71.61 (+47.06) \\
            & + DFT & 36.40 (+6.24) & 38.76 (+14.46) & 52.75 (+17.94) & 64.35 (+17.81) & 23.75 (+13.24) & 82.08 (+57.53) \\
            \cmidrule(l){2-8}
            & \textbf{+ GFT(no mix)} & 42.18 (+12.02) & 39.37 (+15.07) & 55.59 (+20.78) & 68.13 (+21.59) & 27.77 (+17.26) & 82.21 (+57.66) \\
            & \textbf{+ GFT (Ours)} & \cellcolor{blue!10}\textbf{46.09 (+15.93)} & \cellcolor{blue!10}\textbf{40.51 (+16.21)} & \cellcolor{blue!10}\textbf{58.32 (+23.51)} & \cellcolor{blue!10}\textbf{70.50 (+23.96)} & \cellcolor{blue!10}\textbf{28.93 (+18.42)} & \cellcolor{blue!10}\textbf{85.24 (+60.69)} \\
            \midrule
              
            \multirow{9}{*}{\textbf{Qwen2.5-Math-7B}} 
            & Base Model & 42.66 & 34.31 & 49.50 & 59.10 & 19.20 & 85.32 \\
            & + SFT & 41.88 (-0.78) & 38.31 (+4.00) & 54.69 (+5.19) & 67.16 (+8.06) & 31.82 (+12.62) & 87.67 (+2.35) \\
            & + SFT(mix) & 43.06 (+0.40) & 39.47 (+5.16) & 56.83 (+7.33) & 69.63 (+10.53) & 32.45 (+13.25) & 88.93 (+3.61) \\
            & + GRPO & 55.63 (+12.97) & 38.65 (+4.34) & 61.63 (+12.13) & 73.29 (+14.19) & 32.60 (+13.40) & 91.18 (+5.86) \\
            & + ASFT & 52.81 (+10.15) & \cellcolor{blue!10}\textbf{40.76 (+6.45)} & 61.55 (+12.05) & 74.31 (+15.21) & 32.47 (+13.27) & 89.37 (+4.05) \\
            & + PSFT & 41.56 (-1.10) & 38.05 (+3.74) & 56.74 (+7.24) & 67.30 (+8.20) & 34.86 (+15.66) & 83.55 (-1.77) \\
            & + DFT & 51.09 (+8.43) & 39.31 (+5.00) & 57.46 (+7.96) & 70.42 (+11.32) & 35.31 (+16.11) & 86.94 (+1.62) \\
            \cmidrule(l){2-8}
            & \textbf{+ GFT(no mix)} & 53.21 (+10.55) & 38.74 (+4.43) & 61.72 (+12.22) & 74.78 (+15.68) & 34.22 (+15.02) & 92.90 (+7.58) \\
            & \textbf{+ GFT (Ours)} & \cellcolor{blue!10}\textbf{56.09 (+13.43)} & 40.24 (+5.93) & \cellcolor{blue!10}\textbf{63.47 (+13.97)} & \cellcolor{blue!10}\textbf{77.31 (+18.21)} & \cellcolor{blue!10}\textbf{39.86 (+20.66)} & \cellcolor{blue!10}\textbf{93.81 (+8.49)} \\
            \midrule
              
            \multirow{9}{*}{\textbf{DeepSeekMath-7B-Instruct}} 
            & Base Model & 16.09 & 27.56 & 38.00 & 42.73 & 19.44 & 75.70 \\
            & + SFT & 20.93 (+4.84) & 31.53 (+3.97) & 43.13 (+5.13) & 46.51 (+3.78) & 18.71 (-0.73) & 79.30 (+3.60) \\
            & + SFT(mix) & 21.57 (+5.48) & \cellcolor{blue!10}\textbf{32.28 (+4.72)} & 42.43 (+4.43) & \cellcolor{blue!10}\textbf{47.52 (+4.79)} & 20.00 (+0.56) & 78.97 (+3.27) \\
            & + GRPO & 16.72 (+0.63) & 27.59 (+0.03) & 42.18 (+4.18) & 43.39 (+0.66) & 19.39 (-0.05) & 77.74 (+2.04) \\
            & + ASFT & 15.52 (-0.57) & 28.40 (+0.84) & 39.03 (+1.03) & 44.51 (+1.78) & 15.38 (-4.06) & 77.42 (+1.72) \\
            & + PSFT & \cellcolor{blue!10}\textbf{25.78 (+9.69)} & 30.05 (+2.49) & 43.84 (+5.84) & 45.36 (+2.63) & 18.20 (-1.24) & 78.91 (+3.21) \\
            & + DFT & 24.37 (+8.28) & 30.68 (+3.12) & 43.90 (+5.90) & 47.01 (+4.28) & 19.00 (-0.44) & 79.88 (+4.18) \\
            \cmidrule(l){2-8}
            & \textbf{+ GFT(no mix)} & 18.43 (+2.34) & 30.90 (+3.34) & 42.56 (+4.56) & 45.14 (+2.41) & 19.38 (-0.06) & 77.97 (+2.27) \\
            & \textbf{+ GFT (Ours)} & 23.12 (+7.03) & 30.98 (+3.42) & \cellcolor{blue!10}\textbf{48.15 (+10.15)} & 44.79 (+2.06) & \cellcolor{blue!10}\textbf{20.42 (+0.98)} & \cellcolor{blue!10}\textbf{80.08 (+4.38)} \\
            \midrule
              
            \multirow{9}{*}{\textbf{LLaMA-3.2-3B-Instruct}} 
            & Base Model & 23.78 & 25.40 & 38.06 & 44.63 & 14.83 & 69.12 \\
            & + SFT & 19.53 (-4.25) & 26.12 (+0.72) & 36.33 (-1.73) & 43.66 (-0.97) & 12.14 (-2.69) & 68.40 (-0.72) \\
            & + SFT(mix) & 21.09 (-2.69) & 27.88 (+2.48) & 37.76 (-0.30) & 45.68 (+1.05) & 14.15 (-0.68) & 70.00 (+0.88) \\
            & + GRPO & 23.25 (-0.53) & 28.01 (+2.61) & 40.61 (+2.55) & 46.18 (+1.55) & 18.44 (+3.61) & 67.53 (-1.59) \\
            & + ASFT & 18.44 (-5.34) & 26.13 (+0.73) & 37.49 (-0.57) & 43.65 (-0.98) & 11.38 (-3.45) & 66.64 (-2.48) \\
            & + PSFT & 24.37 (+0.59) & 28.94 (+3.54) & 40.73 (+2.67) & 47.43 (+2.80) & 15.41 (+0.58) & 70.91 (+1.79) \\
            & + DFT & 14.21 (-9.57) & 26.63 (+1.23) & 35.13 (-2.93) & 41.45 (-3.18) & 9.58 (-5.25) & 67.39 (-1.73) \\
            \cmidrule(l){2-8}
            & \textbf{+ GFT(no mix)} & \cellcolor{blue!10}\textbf{27.66 (+3.88)} & \cellcolor{blue!10}\textbf{31.14 (+5.74)} & \cellcolor{blue!10}\textbf{43.88 (+5.82)} & \cellcolor{blue!10}\textbf{51.71 (+7.08)} & \cellcolor{blue!10}\textbf{21.29 (+6.46)} & 72.87 (+3.75) \\
            & \textbf{+ GFT (Ours)} & 27.50 (+3.72) & 30.15 (+4.75) & 43.25 (+5.19) & 49.60 (+4.97) & 18.84 (+4.01) & \cellcolor{blue!10}\textbf{72.93 (+3.81)} \\
            \bottomrule
        \end{tabular}%
    }
\end{table*}

Based on Table~\ref{tab:main_results}, GFT demonstrates strong data efficiency under a reduced training budget: with only 10k training examples, it matches or even surpasses a range of baselines trained with 100k examples. Crucially, mixing in distillation data yields only marginal changes for both SFT and GFT (i.e., \textit{SFT(mix)} $\approx$ \textit{SFT} and \textit{GFT(no mix)} $\approx$ \textit{GFT}), indicating that the gains are not primarily driven by additional distilled traces but by the proposed training mechanism. Notably, for smaller heterogeneous models like Llama-3.2-3B, \textsc{GFT (no mix)} surpasses mixing strategies, implying they are less robust to the distribution mismatch from the teacher's distinct reasoning patterns. This superior performance is consistently observed across different model scales and model families, suggesting that the improvements are largely model-agnostic. In terms of the performance profile, GFT yields more uniform gains: whereas some methods exhibit uneven improvements or trade-offs across benchmarks, GFT tends to improve performance across diverse evaluations simultaneously. This suggests that GFT is not merely adapting to a specific question format, but is more reliably improving the quality of the underlying reasoning process. Meanwhile, GRPO can be close to GFT because both are largely driven by GAL that converts sparse (often near-binary) rewards into lower-variance, more informative signals; moreover, under our training setting without explicit KL regularization, GRPO's implicit update stabilization can partially overlap with the effect of our DCR, effectively thereby narrowing down the apparent gap between them.


\subsection{Ablation Studies}
\label{sec:ablation}

\begin{figure}[!t]
    \centering
    \includegraphics[width=0.5\linewidth]{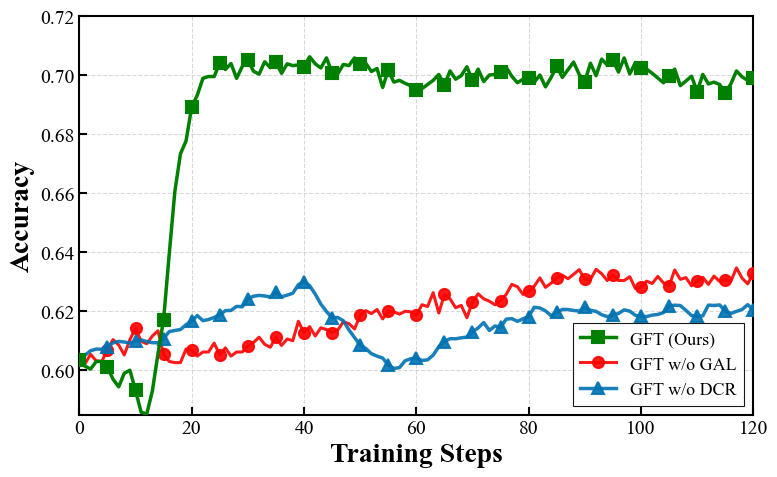}
    \caption{Learning dynamics on MATH-lighteval. Removing DCR causes severe volatility, while removing GAL results in slow convergence and a lower ceiling.}
    \label{fig:abl_dynamic}
\end{figure}

We validate the contributions of GAL and DCR via ablations on Qwen2.5-Math-1.5B, comparing the full GFT with variants that remove GAL, remove DCR, or remove both (equivalent to standard SFT). We report results on Math, AMC23, and Olympiad Bench to cover increasing difficulty and robustness requirements, and further inspect the optimization behavior of each variant using the learning-dynamics plot in Figure \ref{fig:abl_dynamic}.

\begin{table}[!t]
    \centering
    \footnotesize
    \caption{Ablation on \textbf{Qwen2.5-Math-1.5B}. GAL is important for complex reasoning (e.g., Olympiad) and DCR enhances performance by ensuring optimization stability, their synergy yields optimal results.}
    \label{tab:ablation}
    
    \setlength{\tabcolsep}{4pt}
    \renewcommand{\arraystretch}{1.2}
    
    \begin{tabular}{lccc}
        \toprule
        \textbf{Method} & \textbf{AMC23} & \textbf{MATH} & \textbf{Olympiad} \\
        \midrule
        Base Model & 30.16 & 46.54 & 23.39 \\
        \midrule
        GFT w/o (GAL + DCR) & 31.25 & 60.66 & 24.58 \\
        GFT w/o GAL & 35.78 & 63.91 & 26.63 \\
        GFT w/o DCR & 42.81 & 65.97 & 27.82 \\
        \textbf{GFT (Ours)} & \textbf{46.09} & \textbf{70.50} & \textbf{30.52} \\
        \bottomrule
    \end{tabular}
\end{table}

The results in Table \ref{tab:ablation} demonstrate the distinct contributions of each component.
Removing GAL causes the sharpest decline on the hardest benchmark (Olympiad), validating that group-based contrastive feedback is vital for extracting signals in complex reasoning.
In contrast, removing DCR primarily impacts robustness (Minerva), consistent with its role in rectifying gradient explosion.
These performance patterns are further corroborated by the learning dynamics in Figure \ref{fig:abl_dynamic}: the removal of DCR leads to severe training volatility, while removing GAL results in slow, suboptimal convergence.
Ultimately, GFT synergizes both components to ensure efficient and stable optimization.

\subsection{Compatibility with SFT and RL}
\label{sec:compatibility}



\begin{figure*}[!t]
    \centering
    \includegraphics[width=1\linewidth]{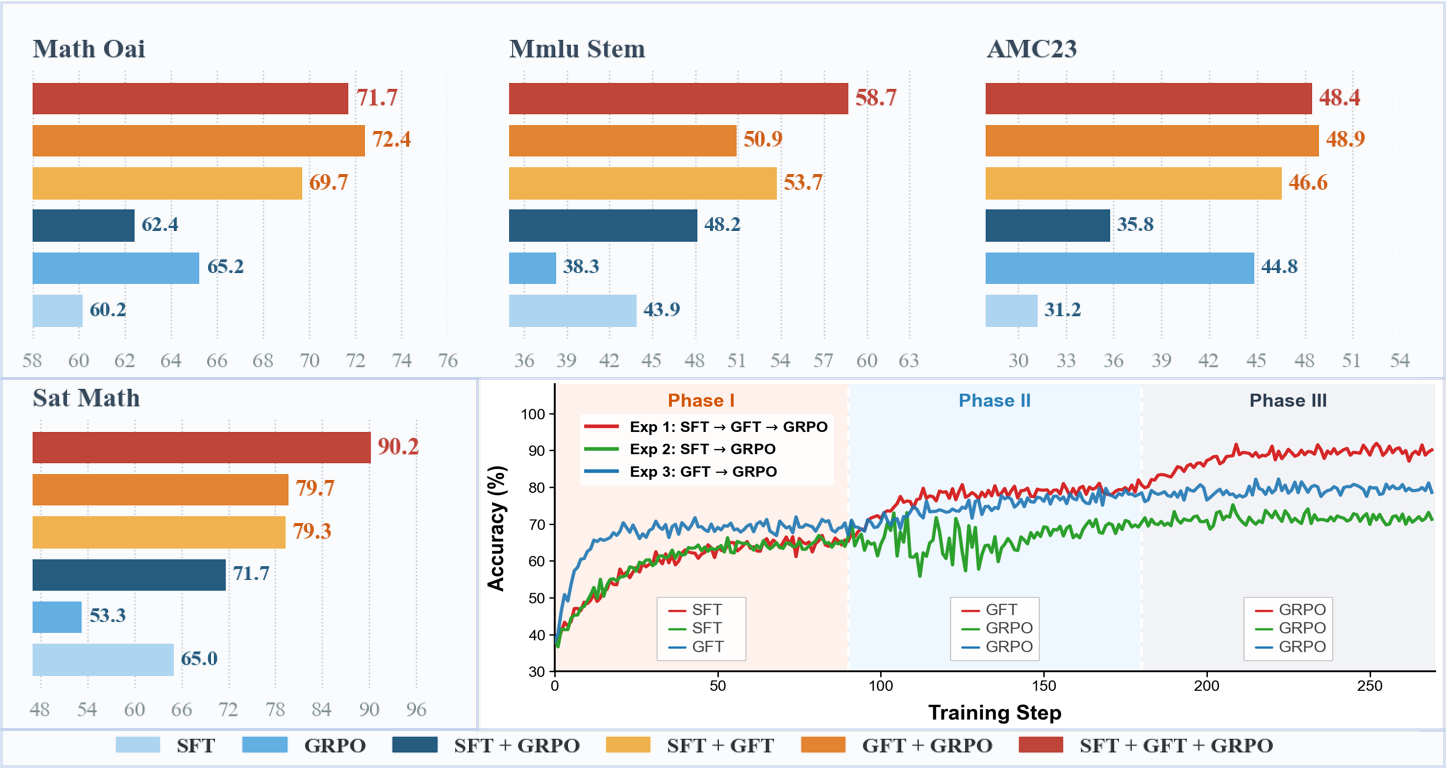}
    \caption{Performance comparison on Qwen2.5-Math-1.5B (Pass@16). Bottom-right: Sat-Math training dynamics. SFT+GFT+GRPO achieves top performance via stable optimization, demonstrating GFT's high compatibility and effective synergy between SFT and GRPO.}
    \label{fig:Compatibility_of_GFT}
\end{figure*}

We conduct a sequential-training compatibility study by combining SFT, GFT, and GRPO in different compositions (Figure~\ref{fig:Compatibility_of_GFT}). This design aims to diagnose the \emph{synergy dilemma} in conventional post-training—where SFT may rigidify the policy and narrow the effective exploration manifold for downstream RL—and to evaluate whether GFT can both (i) serve as a stronger initializer for RL and (ii) improve the handoff from SFT to RL.

As shown in Figure~\ref{fig:Compatibility_of_GFT}, we design GFT to improve compatibility in two aspects. \textbf{(1) To improve RL exploration,} GAL prevents the cold-start policy from collapsing to a single expert-induced mode and maintains a multi-solution distribution via group-wise relative advantages. This broader support produces more diverse rollouts and stronger advantage signals for GRPO, explaining why \textit{GFT + GRPO} gains more than \textit{SFT + GRPO} on harder benchmarks \cite{li2024preserving}. \textbf{(2) To prevent distribution extremization and preserve exploration,} DCR bounds per-token updates to avoid over-sharpening an SFT-initialized policy. Without this constraint, large steps can quickly drive the policy to a low-entropy, mode-concentrated distribution, reducing rollout diversity and weakening GRPO’s learning signal. By limiting update magnitude, DCR keeps the policy in a higher-entropy regime, matching the smoother dynamics and higher ceiling of \textit{SFT + GFT + GRPO} in Figure~\ref{fig:Compatibility_of_GFT}. \textbf{Notably,} \textit{GFT + GRPO} surpassing \textit{SFT + GRPO} does not mean GFT replaces SFT: SFT provides a reliable initialization point for alignment and formatting, while GFT improves RL compatibility by preserving support and stabilizing updates. Thus, \textit{SFT + GFT + GRPO} works best as a staged pipeline: SFT sets the initialization point, GFT restores exploration capabilities without drifting, and GRPO leverages higher-quality trajectories to reach the top ceiling.




\subsection{Catastrophic Forgetting Analysis}
\label{sec:analysis_kl}

\begin{table}[!t]
    \centering
    \footnotesize
   \caption{Performance of ~\textbf{LLaMA-3.2-3B-Instruct} on general reasoning benchmarks. While SFT induces substantial catastrophic forgetting, GFT largely preserves base performance.}
    \label{tab:forgetting_analysis}
    
    \begin{minipage}{0.6\linewidth}
    \centering
    \renewcommand{\arraystretch}{1.1}
    \setlength{\tabcolsep}{2pt}
    
    \begin{tabular*}{\linewidth}{@{\extracolsep{\fill}}lccc}
        \toprule
        \textbf{Method} & \textbf{Mawps} & \textbf{Svamp} & \textbf{Mmlu stem} \\ 
        \midrule
        Base Model & 96.06 & 86.36 & 41.03 \\
        \midrule
        +SFT & 91.97 (- 4.09) & 78.73 (- 7.63) & 35.05 (-5.98) \\
        
        +GRPO & 94.60 (-\phantom{0}1.46) & \textbf{88.11 (+\phantom{0}1.75)} & 39.48 (-1.55) \\
        
        +GFT (Ours) & \textbf{95.79 (-\phantom{0}0.27)} & 84.65 (-\phantom{0}1.71) & \textbf{43.89 (+2.86)} \\
        \bottomrule
    \end{tabular*}
    \end{minipage}
\end{table}

\begin{figure}[!t]
    \centering
    \includegraphics[width=0.5\linewidth]{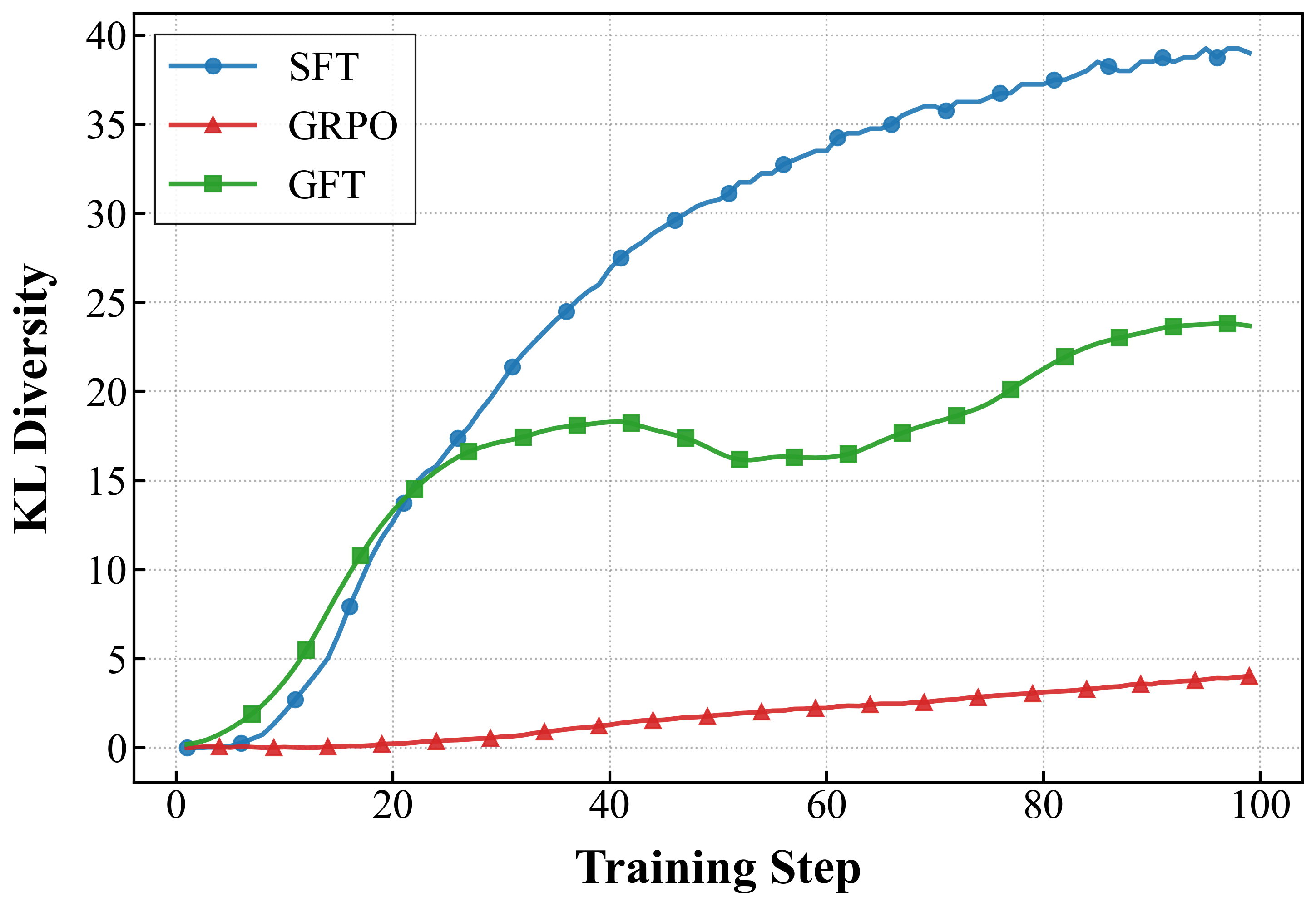}
    \caption{KL divergence quantifies distributional drift from the base model. SFT exhibits the highest divergence, while GFT maintains a significantly lower level, effectively mitigating catastrophic forgetting.}
    \label{fig:generation}
\end{figure}

Table~\ref{tab:forgetting_analysis} shows a clear contrast in catastrophic forgetting on general reasoning benchmarks.
After domain training, \textbf{SFT} exhibits substantial degradation on \textsc{MAWPS} and \textsc{SVAMP} and also drops on \textsc{MMLU-STEM}, indicating severe forgetting.
In contrast, \textbf{GRPO} largely preserves the base model's prior capabilities, while \textbf{GFT} not only maintains comparable retention to GRPO but also improves \textsc{MMLU-STEM}.
This ranking is further consistent with Figure~\ref{fig:generation}, where the policy shift of SFT is the most pronounced, whereas GRPO and GFT remain significantly closer to the base policy.

To quantify forgetting more directly, we adopt the approach of~\cite{shenfeld2025rl} and compute the \emph{average KL divergence} between the trained model and the base model on the training dataset. Recent empirical studies further support the correlation between this KL-based drift and forgetting~\citep{chu2025sft, luo2025empirical, ruan2025unveiling}. We therefore use the average KL divergence as a proxy for distributional drift, and hence forgetting.
We analyze the training dynamics of Qwen2.5-Math-1.5B across different methpds.
As shown in Figure~\ref{fig:generation}, all baselines converge to their peak performance approximately at step 100.
At this stage, we observe a distinct contrast: \textbf{SFT} incurs the highest alignment tax with the largest KL divergence, whereas \textbf{GRPO} retains a \emph{KL-minimal} solution; notably, \textbf{GFT} strikes a balance, stabilizing at a low KL level comparable to GRPO.
We attribute this stability to our design: \textbf{GAL} reinforces high-quality output trajectories in a reward-driven manner, avoiding abrupt distributional shifts induced by pure cross-entropy trace fitting; meanwhile, \textbf{DCR} suppresses gradient explosions from ``extreme tokens'' (where $\pi_{\theta}\!\approx\!0$), preventing drastic policy drift.
Together, these components enable efficient knowledge injection while retaining robust general-purpose reasoning.

\subsection{Diversity of GFT}
\label{sec:analysis_diversity}

Balancing solution diversity with correctness remains a challenge in post-training. While distillation preserves exploration by mimicking the teacher's soft targets~\cite{goyal2025distilled}, it often lacks explicit correctness incentives. Conversely, RL-style optimization (e.g., GRPO) tends to sharpen the policy toward specific high-reward trajectories, which effectively optimizes precision but may suppress the exploration space and reduce solution variety~\cite{yue2025does}. To evaluate whether GFT can effectively reconcile this trade-off—maintaining intrinsic diversity while ensuring accuracy—we conduct a multi-sample evaluation using Pass@$k$ as a proxy metric for solution coverage. Table~\ref{tab:passk} compares the diversity performance of \textbf{Distillation}, \textbf{GRPO}, and \textbf{GFT}.



\begin{table}[!t]
    \centering
    \footnotesize
    \caption{Comparison of Pass@k ($k=128, 256$) performance between Distillation, GRPO and GFT. GFT consistently achieves the highest Pass@k scores, effectively enhancing response diversity.}
    \label{tab:passk}
    
    \begin{minipage}{0.5\linewidth}
    \centering
    \setlength{\tabcolsep}{2pt}
    \renewcommand{\arraystretch}{1.2}
    
    \begin{tabular*}{\linewidth}{@{\extracolsep{\fill}}c|c|cccc}
        \toprule
        \textbf{Metric} & \textbf{Method} & \textbf{SAT Math} & \textbf{Minerva} & \textbf{TabMWP} & \textbf{Avg.} \\
        \midrule
        \multirow{4}{*}{Pass@128} 
        & Base Model   & 39.69 & 9.71  & 24.17 & 24.52 \\
        & Distillation & 66.67 & 22.98 & 79.32 & 56.32 \\
        & GRPO         & 52.95 & 19.89 & 76.77 & 49.87 \\
        & \textbf{GFT} & \textbf{72.58} & \textbf{28.59} & \textbf{85.31} & \textbf{62.16} \\
        \midrule
        \multirow{4}{*}{Pass@256} 
        & Base Model   & 38.76 & 9.25  & 24.36 & 24.12 \\
         & Distillation & 67.20 & 21.84 & 79.28 & 56.11 \\
          & GRPO         & 51.90 & 19.77 & 75.82 & 49.16 \\
          & \textbf{GFT} & \textbf{73.33} & \textbf{27.17} & \textbf{85.23} & \textbf{61.91} \\
          \bottomrule
    \end{tabular*}
    \end{minipage}
\end{table}

GFT achieves the highest Pass@128 and Pass@256 across benchmarks. Distillation improves exploration because soft targets from teacher train the student to match the teacher’s \emph{output distribution}, but it does not use reward to distinguish correct reasoning. GRPO, in contrast, uses reward to \emph{sharpen} the student distribution, which strengthens memory of rewarded (often correct) paths but also narrows exploration. GFT combines both signals by reward-evaluating trajectories from \emph{both} the teacher distribution and the student’s own sampling distribution: it learns the teacher’s diverse modes (as in distillation) while using within-group advantages to explicitly compare student samples against teacher traces, pushing the student toward the teacher’s \emph{high-reward diverse} modes. This teacher--student gap correction preserves diversity where it matters, leading to higher Pass@$k$.


\subsection{Hyperparameter Analysis}
\label{sec:hyperparameter}



\begin{table}[!t]
    \centering
    \footnotesize
    \caption{Impact of group composition ratio ($N_{demo}:N_{sample}$); \textbf{2:6} achieves the best accuracy, indicating richer contrast from self-samples with demo samples.}
    \label{tab:hyper_ratio}
      
    \begin{minipage}{0.5\linewidth}
    \centering
    \setlength{\tabcolsep}{3pt}
    \renewcommand{\arraystretch}{1.2}
      
    \begin{tabular*}{\linewidth}{@{\extracolsep{\fill}}ccccc}
        \toprule
        \textbf{Ratio} & \textbf{Minerva Math} & \textbf{Olympiad} & \textbf{Sat Math} & \textbf{Avg.} \\
        \midrule
        8 : 0 & 15.11 & 22.48 & 36.92 & 24.84 \\
        6 : 2 & 29.53 & 29.60 & 71.68 & 43.60 \\
        4 : 4 & 28.93 & 30.52 & 69.93 & 43.13 \\
        \textbf{2 : 6} & \textbf{31.01} & \textbf{32.73} & \textbf{73.04} & \textbf{45.59} \\
        0 : 8 & 23.31 & 28.61 & 40.60 & 30.84 \\
        \bottomrule
    \end{tabular*}
    \end{minipage}
\end{table}

\begin{figure}[!t]
    \centering
    \includegraphics[width=0.5\linewidth]{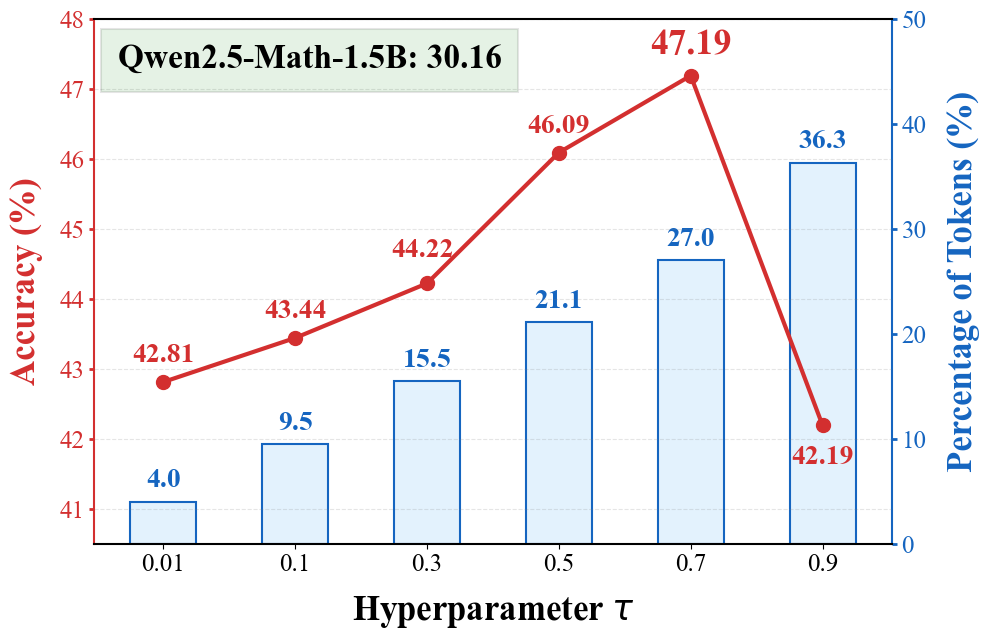}
    \caption{Effect of the clipping threshold $\tau$: larger $\tau$ rectifies more tokens. Accuracy follows an inverted U-shape; insufficient clipping is unstable, while excessive clipping reduces learning efficiency.}
    \label{fig:hyper_tau}
\end{figure}


To probe the impact of \textbf{group diversity} and \textbf{rectification strength}, we ablate the composition ratio ($N_{\mathrm{demo}}:N_{\mathrm{sample}}$) and threshold $\tau$ on Qwen2.5-Math-1.5B.
With fixed $K=8$, Table~\ref{tab:hyper_ratio} identifies \textbf{2:6} as optimal, where minimal demonstrations \textit{anchor} correctness while abundant self-samples provide richer \textit{contrastive signals} for advantage learning.
Regarding the clipping threshold $\tau$, Figure~\ref{fig:hyper_tau} reports accuracy together with the fraction of DCR-rectified tokens. As $\tau$ increases, the rectification rate rises monotonically, indicating stronger clipping. Meanwhile, accuracy exhibits an \textbf{inverted U-shape}: small $\tau$ yields insufficient clipping and unstable updates, whereas large $\tau$ over-clips many tokens and attenuates informative gradients, harming learning efficiency. Consequently, $\tau \approx 0.7$ achieves the best \textbf{stability--efficiency trade-off} in learning. Notably, GFT consistently outperforms the base model across the entire sweep of parameters, suggesting that DCR is robust to $\tau$.


\section{Conclusion}
In this work, we analyze SFT as a special case of RL. This perspective reveals two intrinsic limitations: single-path dependency that restricts exploration, and gradient explosion that causes instability. To address these, we propose \textbf{Group Fine-Tuning (GFT)}. This framework leverages Group Advantage Learning to enhance diversity via contrastive supervision and employs Dynamic Coefficient Rectification to stabilize optimization by preventing extreme weight updates. Experiments demonstrate that GFT effectively balances efficient knowledge injection with robust generalization, offering a principled paradigm for post-training.

\section{Limitations}
Despite GFT's effectiveness, we acknowledge three limitations. First, our evaluation focuses on mathematical reasoning with objective correctness; extending GFT to open-ended tasks with subjective rewards requires further exploration. Second, constructing response groups introduces marginal data preparation overhead compared to standard SFT, though this cost is significantly lower than online RL. Third, due to academic resource constraints, our experiments are limited to models up to 8B parameters; validating GFT on 70B+ models remains an important future direction.

\newpage
\section*{Acknowledgments}
This work is supported by the Key R\&D Program of Ningbo under Grant No.2024Z115


\bibliographystyle{assets/acl_natbib}
\bibliography{paper}

\clearpage
\beginappendix
\section{Derivation: Viewing SFT as a Special Case of On-Policy RL}
\label{sec:proof}

In this appendix, we provide a detailed derivation showing that supervised fine-tuning (SFT) can be interpreted as a special case of reinforcement learning (RL) with a sparse reward function. Specifically, we show that the gradient of the SFT objective can be rewritten as an on-policy expectation under the current policy via importance sampling.

\subsection{SFT Objective and Gradient}

We consider a dataset of expert demonstrations
\(
\mathcal{D} = \{(x, y^*)\}
\),
where \(x\) denotes the input and \(y^*\) is the expert-provided output. The standard SFT objective is defined as the negative log-likelihood:
\begin{equation}
\mathcal{L}_{\mathrm{SFT}}(\theta)
=
-\mathbb{E}_{(x,y^*) \sim \mathcal{D}}
\left[
\log \pi_\theta(y^* \mid x)
\right].
\end{equation}

Taking the gradient with respect to the model parameters \(\theta\), we obtain
\begin{equation}
\nabla_\theta \mathcal{L}_{\mathrm{SFT}}(\theta)
=
-
\mathbb{E}_{(x,y^*) \sim \mathcal{D}}
\left[
\nabla_\theta \log \pi_\theta(y^* \mid x)
\right].
\label{eq:sft_grad_expert}
\end{equation}

This expectation is taken over the expert data distribution rather than samples generated by the current policy.

\subsection{Importance Sampling Reformulation}

We factorize the expert data distribution as
\begin{equation}
P(x, y^*) = P(x)\, P_{\mathrm{expert}}(y^* \mid x),
\end{equation}
and define the joint distribution induced by the current policy as
\begin{equation}
Q(x, y) = P(x)\, \pi_\theta(y \mid x).
\end{equation}

Since both distributions share the same marginal \(P(x)\), we can apply importance sampling to rewrite the expectation in Eq.~\eqref{eq:sft_grad_expert} under \(Q(x,y)\):
\begin{equation}
\begin{aligned}
&\nabla_\theta \mathcal{L}_{\mathrm{SFT}}(\theta)\\
&=
-
\mathbb{E}_{(x,y) \sim Q}
\left[
\nabla_\theta \log \pi_\theta(y \mid x)
\cdot
\frac{P_{\mathrm{expert}}(y \mid x)}{\pi_\theta(y \mid x)}
\right].
\end{aligned}
\label{eq:sft_importance}
\end{equation}

For deterministic expert demonstrations, the expert conditional distribution reduces to a Dirac delta:
\begin{equation}
P_{\mathrm{expert}}(y \mid x) = \mathbb{I}[y = y^*].
\end{equation}

Substituting this into Eq.~\eqref{eq:sft_importance} yields
\begin{equation}
\begin{aligned}
&\nabla_\theta \mathcal{L}_{\mathrm{SFT}}(\theta)\\
&=
-
\mathbb{E}_{(x,y) \sim Q}
\left[
\frac{\mathbb{I}[y = y^*]}{\pi_\theta(y \mid x)}
\nabla_\theta \log \pi_\theta(y \mid x)
\right].
\end{aligned}
\label{eq:sft_final_form_app}
\end{equation}

This recovers the equivalent on-policy formulation presented in the main text.

\subsection{Reinforcement Learning Interpretation}

Equation~\eqref{eq:sft_final_form_app} admits a direct reinforcement learning interpretation. In particular, it corresponds to an on-policy policy gradient with:
\begin{itemize}
    \item \textbf{Policy:} \(\pi_\theta(y \mid x)\);
    \item \textbf{Reward function:}
    \begin{equation}
    r(x,y) = \mathbb{I}[y = y^*],
    \end{equation}
    which provides a unit reward only when the sampled output exactly matches the expert demonstration;
    \item \textbf{Importance weight:}
    \begin{equation}
    w(x,y) = \frac{1}{\pi_\theta(y \mid x)},
    \end{equation}
    correcting for sampling from the model policy instead of the expert distribution.
\end{itemize}

Under this view, SFT can be regarded as a degenerate RL setting with an extremely sparse reward signal and high variance, where learning occurs only through trajectories that coincide exactly with expert demonstrations.

\subsection{Summary}

In summary, the derivation proceeds by (i) expressing the SFT gradient as an expectation over expert data, (ii) applying importance sampling to rewrite it under the model policy, and (iii) specializing the expert distribution to a deterministic form. This establishes a formal equivalence between SFT and on-policy reinforcement learning with a sparse indicator reward, providing a unified perspective on supervised and reinforcement-based post-training.

\section{Formulation of Group Fine-Tuning}
\label{sec:formulation}

In this appendix, we provide the explicit loss formulations and gradient expressions of Group Fine-Tuning (GFT), including both sequence-level and token-level forms. These formulations correspond to the gradient expression presented in Eq.~\eqref{eq:gft_gradient} in the main text.

\subsection{Sequence-Level Objective}
\label{subsec:seq_level_obj}

For each input query \(x\), we construct a response group
\(
\mathcal{G}_x = \{y_1, \ldots, y_K\}
\),
where each response \(y_k\) is assigned a scalar reward \(R(y_k)\) and a standardized
group advantage \(A(y_k)\) as defined in Eq.~\eqref{eq:advantage}. We define the sequence-level GFT loss as
\begin{equation}
\begin{aligned}
\mathcal{L}_{\mathrm{GFT}}^{\mathrm{seq}}(\theta)
=
-
\mathbb{E}_{x}
\Bigg[
\sum_{y_k \in \mathcal{G}_x}
A(y_k)\,
\mathcal{C}\!\left(\pi_\theta(y_k \mid x)\right)
\\[-1mm]
\quad\cdot
\log \pi_\theta(y_k \mid x)
\Bigg].
\end{aligned}
\label{eq:gft_loss_seq}
\end{equation}

where \(\mathcal{C}(\cdot)\) is the dynamic coefficient rectification function defined
in Eq.~\eqref{eq:rectification_factor}.

Taking the gradient of Eq.~\eqref{eq:gft_loss_seq} yields the sequence-level policy gradient:
\begin{equation}
\begin{split}
\nabla_\theta \mathcal{L}_{\mathrm{GFT}}^{\mathrm{seq}}
&= 
\mathbb{E}_{x}
\Bigg[
\sum_{y_k \in \mathcal{G}_x}
A(y_k)\,
\frac{\mathcal{C}\left(\pi_\theta(y_k \mid x)\right)}{\pi_\theta(y_k \mid x)}
\\
&\qquad\cdot 
\nabla_\theta \log \pi_\theta(y_k \mid x)
\Bigg].
\end{split}
\label{eq:gft_grad_seq}
\end{equation}

\subsection{Token-Level Decomposition}
\label{subsec:token_level_decomp}

Each response sequence \(y_k = (y_{k,1}, \ldots, y_{k,T_k})\) is generated autoregressively by the policy:
\begin{equation}
\pi_\theta(y_k \mid x)
=
\prod_{t=1}^{T_k}
\pi_\theta\!\left(y_{k,t} \mid y_{k,<t}, x\right).
\end{equation}

Accordingly, the sequence log-probability decomposes as
\begin{equation}
\log \pi_\theta(y_k \mid x)
=
\sum_{t=1}^{T_k}
\log \pi_\theta\!\left(y_{k,t} \mid y_{k,<t}, x\right).
\end{equation}

We use the shorthand
\begin{equation}
\pi_{k,t}
\triangleq
\pi_\theta\!\left(y_{k,t} \mid y_{k,<t}, x\right).
\label{eq:pi_kt_def}
\end{equation}
for the token-level prediction probability.

Substituting the above decomposition into Eq.~\eqref{eq:gft_loss_seq}, we obtain the token-level GFT loss:
\begin{equation}
\begin{split}
& \mathcal{L}_{\mathrm{GFT}}^{\mathrm{tok}}(\theta)
= 
\\
& - 
\mathbb{E}_{x}
\Bigg[
\sum_{y_k \in \mathcal{G}_x}
A(y_k)\,
\sum_{t=1}^{T_k}
\mathcal{C}(\pi_{k,t})\,
\log \pi_{k,t}
\Bigg].
\end{split}
\label{eq:gft_loss_tok}
\end{equation}

Taking the gradient yields the token-level policy gradient:
\begin{equation}
\begin{split}
& \nabla_\theta \mathcal{L}_{\mathrm{GFT}}^{\mathrm{tok}}
= 
\\
& \mathbb{E}_{x} 
\Bigg[
\sum_{y_k \in \mathcal{G}_x}
A(y_k)\,
\sum_{t=1}^{T_k}
\frac{\mathcal{C}(\pi_{k,t})}{\pi_{k,t}}\,
\nabla_\theta
\log \pi_{k,t}
\Bigg].
\end{split}
\label{eq:gft_grad_tok}
\end{equation}

\subsection{Relation to SFT and RL Objectives}
\label{subsec:relation_sft_rl}

When the response group degenerates to a single expert demonstration (\(|\mathcal{G}_x| = 1\)), the advantage is constant and Eq.~\eqref{eq:gft_grad_tok} reduces to the standard SFT gradient. Conversely, when the group consists of diverse sampled trajectories with non-trivial advantage values, GFT recovers an on-policy reinforcement learning update with group-normalized advantage weighting and bounded importance coefficients.

This formulation establishes GFT as a strict generalization of SFT and a stabilized, contrastive variant of policy-gradient-based post-training.

\section{Evaluation Settings}
\label{sec:eval_settings}
We conduct evaluations on a broad suite of 11 benchmarks: AMC23~\citep{amc2023dataset}, College Math~\citep{hendrycks2020measuring}, Gaokao~\citep{zhang2023evaluating}, Math~\citep{hendrycks2021measuring}, Minerva Math~\citep{lewkowycz2022solving}, TabMWP~\citep{lu2022dynamic}, OlympiadBench~\citep{he-etal-2024-olympiadbench}, Mmlu Stem~\citep{hendrycks2020measuring}, Sat Math~\citep{zhong2024agieval}, Mawps~\citep{koncel2016mawps}, and Svamp~\citep{patel2021nlp}. These benchmarks are carefully selected to cover a wide spectrum of difficulty levels and reasoning types, ensuring a holistic assessment of the model's capabilities. We report the average Pass@1 accuracy across 16 decoding runs (Pass@16 Average) with a sampling temperature of 0.5 and a maximum generation length of 4096 tokens.

\paragraph{Trade-off Between SFT and RL}
Post-training paradigms typically navigate a trade-off between Supervised Fine-Tuning (SFT) and Reinforcement Learning (RL). SFT is widely recognized for its efficiency in knowledge injection and ``cold-starting''~\citep{zhou2023lima, chung2024scaling}; however, it is prone to mechanical memorization and often fails to generalize to out-of-distribution scenarios~\citep{ouyang2022training, bai2022training, chu2025sft, swamy2025all, huan2025does}. Conversely, RL excels at discovering robust strategies and optimizing long-term objectives~\citep{christiano2017deep}, yet it is computationally expensive and struggles to acquire complex reasoning skills from scratch without sufficient guidance~\citep{schulman2017proximal, sheng2025hybridflow, mandlekar2021matters, chen2025empirical}.

\paragraph{The Synergy Dilemma in Hybrid Post-Training}
Standard hybrid approaches (e.g., SFT followed by RL) attempt to combine these complementary strengths but face a severe ``synergy dilemma''~\citep{ouyang2022training, rafailov2023direct}. Recent studies conclude that this conflict arises from the fundamental training dynamics: the overfitting induced by SFT creates a rigid policy that severely constrains the exploration space required for subsequent RL~\citep{chen2025sft}, while simultaneously leading to reasoning pattern mismatches that hinder effective policy alignment~\citep{chen2025synergy}. Although methods like interleaved updates~\citep{liu2025uft} or preference optimization~\citep{rafailov2023direct} offer partial solutions, they remain dependent on external feedback signals. In contrast, our work addresses this dilemma by transforming the rigid imitation objective into a \textbf{Group Advantage Learning} framework, which explicitly preserves solution diversity and the exploration manifold by optimizing contrastive advantages derived from hybrid response groups.

\paragraph{Single-Stage Hybrids: Mixing Imitation and Exploration}
Several recent studies have attempted to unify SFT and RL by balancing imitation and exploration through modified objectives~\citep{yuan2024self}. Single-stage hybrid methods, such as SRFT~\citep{fu2025srft} and UFT~\citep{liu2025uft}, employ dynamic weighting mechanisms, interleaved updates, or dense verification signals~\citep{wang2024math, yu2024ovm} to mix supervised signals with reinforcement objectives. Similarly, frameworks like HybridFlow~\citep{sheng2025hybridflow} explore flexible combinations of offline and online data to bridge the gap. While approaches like CHORD~\citep{zhu2025anchored} introduce anchor-based constraints to maintain stability, a common limitation across these methods is that they often treat SFT and RL as separate components to be linearly combined or alternated, rather than fusing them mathematically into a cohesive formulation derived from a unified training dynamic.

\paragraph{Gradient-Level Stabilization and Its New Trade-offs}
To address the instability inherent in post-training, other researchers have revisited the underlying gradient formulation. Theoretical analyses suggest a deeper equivalence between likelihood maximization and reinforcement learning~\citep{swamy2025all}, prompting new rectification strategies. For instance, \citet{wu2025generalization} propose Dynamic Fine-Tuning (DFT), which counteracts gradient explosion by reweighting the loss with the model's likelihood to cancel the inverse-probability term. However, this indiscriminate dampening creates a new dilemma: it suppresses the strong gradient signals required for injecting novel knowledge, potentially hindering adaptation to new domains. Alternatively, approaches like Proximal SFT~\citep{zhu2025proximal} and Anchored SFT~\citep{zhu2025anchored} introduce trust-region constraints to stabilize fine-tuning, yet such rigid regularizations may overly constrain the model's plasticity. In the realm of Reinforcement Learning, stability is traditionally enforced via KL-divergence penalties~\citep{ouyang2022training} or clipping mechanisms~\citep{schulman2017proximal}. More recently, group-based methods like GRPO~\citep{shao2024deepseekmath} have emerged to mitigate gradient variance by normalizing advantages within generated groups, effectively removing the reliance on unstable critic models, while system-level frameworks like HybridFlow~\citep{sheng2025hybridflow} attempt to stabilize training through flexible data scheduling.

\end{document}